\documentclass[sigconf]{acmart}

\AtBeginDocument{%
  \providecommand\BibTeX{{%
    \normalfont B\kern-0.5em{\scshape i\kern-0.25em b}\kern-0.8em\TeX}}}

\acmConference[KDD '20 Humanitarian Mapping Workshop]{KDD '20: ACM SIGKDD Conference on Knowledge Discovery and Data Mining (KDD) Humanitarian Mapping Workshop}{August 24, 2020}{San Diego, CA}
\acmBooktitle{KDD '20: ACM SIGKDD Conference on Knowledge Discovery and Data Mining (KDD) Humanitarian Mapping Workshop}

\usepackage{bm}

\begin{document}

\title{Rapid Response Crop Maps in Data Sparse Regions}

\author{Hannah Kerner}
\authornote{Both authors contributed equally to this research.}
\email{hkerner@umd.edu}
\orcid{0000-0002-3259-7759}
\affiliation{%
  \institution{University of Maryland, College Park}
}
\author{Gabriel Tseng}
\authornotemark[1]
\email{gabrieltseng95@gmail.com}
\affiliation{%
  \institution{NASA Harvest}
}
\author{Inbal Becker-Reshef}
\email{ireshef@umd.edu}
\affiliation{%
  \institution{University of Maryland, College Park}
}
\author{Catherine Nakalembe}
\email{cnakalem@umd.edu}
\affiliation{%
  \institution{University of Maryland, College Park}
}
\author{Brian Barker}
\email{bbarker1@umd.edu}
\affiliation{%
  \institution{University of Maryland, College Park}
}
\author{Blake Munshell}
\email{bmunshel@terpmail.umd.edu}
\affiliation{%
  \institution{University of Maryland, College Park}
}
\author{Madhava Paliyam}
\email{mpaliyam@terpmail.umd.edu}
\affiliation{%
  \institution{University of Maryland, College Park}
}
\author{Mehdi Hosseini}
\email{mhoseini@umd.edu}
\affiliation{%
  \institution{University of Maryland, College Park}
}
%
\renewcommand{\shortauthors}{Kerner and Tseng, et al.}

\begin{abstract}
Spatial information on cropland distribution, often called cropland or crop maps, are critical inputs for a wide range of agriculture and food security analyses and decisions. However, high-resolution cropland maps are not readily available for most countries, especially in regions dominated by smallholder farming (e.g., sub-Saharan Africa). These maps are especially critical in times of crisis when decision makers need to rapidly design and enact agriculture-related policies and mitigation strategies, including providing humanitarian assistance, dispersing targeted aid, or boosting productivity for farmers. A major challenge for developing crop maps is that many regions do not have readily accessible ground truth data on croplands necessary for training and validating predictive models, and field campaigns are not feasible for collecting labels for rapid response. We present a method for rapid mapping of croplands in regions where little to no ground data is available. We present results for this method in Togo, where we delivered a high-resolution (10 m) cropland map in under 10 days to facilitate rapid response to the COVID-19 pandemic by the Togolese government. This demonstrated a successful transition of machine learning applications research to operational rapid response in a real humanitarian crisis. All maps, data, and code are publicly available to enable future research and operational systems in data-sparse regions.
\end{abstract}




\maketitle

\section{Introduction}

Satellite Earth observation (EO) data can provide critical, relevant, and timely information in support of agricultural monitoring and food security \citep{BeckerReshef2019,NationalAcademies2019}. EO can enable crop yield assessment and forecasting \citep{Becker2010,Franch2015,Franch2017,Lobell2013,Porter2017}, inform agricultural commodities markets, enable early warning and mitigation of impending crop shortages, and inform decisions about subsidies in times of crisis such as extreme drought \citep{Dorward2011} or outbreaks of disease or pests. However, the lack of up-to-date, high-resolution cropland and crop type maps is a critical barrier to agricultural and food security assessments and operational rapid response to humanitarian crises. These data are especially lacking in regions dominated by smallholder agriculture that are most vulnerable to food insecurity (e.g., sub-saharan Africa). 

Machine learning techniques have commonly been used to identify cropland in EO data. The majority of studies use tree-based classifiers (primarily random forests or decision trees) or neural network/deep learning methods (primarily recurrent or convolutional neural networks). However, prior methods have typically been applied to small and/or spatially homogeneous areas \citep{RNN_Camargue, RNN_winter, Skakun2019}, require large training datasets (hundreds of thousands of labels) \citep{breizhcrops2020,Wang2020,lstm_crop_identification, RF_france}, or use EO inputs with insufficient spatial resolution for reliably detecting smallholder farms which often have field sizes $<$1 ha \citep{GFSAD,Teluguntla2018,Oliphant2019,CopernicusLandCover,Samasse2020,Waldner2017}. In many regions, particularly developing countries and regions dominated by smallholder farming, the lack of readily accessible ground truth data is a major barrier for training and validating machine learning techniques for cropland classification \citep{Coutu2020}. 

In this paper, we demonstrate a novel method for rapidly generating cropland maps for areas with little to no available labeled data over a large heterogeneous area. This is achieved by leveraging global and local datasets. The global dataset contains tens of thousands of crowdsourced labels from diverse geographies \citep{GeoWiki} to learn general features for identifying cropland globally. These global examples supplement a much smaller dataset of curated, local examples (manually labeled using photointerpretation of high-resolution satellite imagery) to learn features specific to cropland in the target mapping region. Combined, these datasets provide sufficient examples to train a Long Short Term Memory network (LSTM) \citep{LSTM} to predict the presence of cropland in each pixel given a monthly time-series of Sentinel-2 multispectral observations. 

The COVID-19 global pandemic has severely impacted the global food system, and the threat of food shortages and rising food prices has required countries at risk of food insecurity to rapidly design and enact aid programs in response. In the West African country of Togo, the government announced a program to boost national food production in response to the COVID-19 crisis by distributing aid to farmers, but found that high-resolution spatial information about the distribution of farms across Togo (which primarily consist of smallholder farms under 1 ha) was a critical piece of missing information for designing this program. Using our method, we were able to generate an accurate high-resolution (10 m) map of cropland in Togo for 2019 in under 10 days to respond to the Togolese government's immediate need for information about the distribution of farms across the country. We showed that our model outperforms existing publicly-available cropland maps for Togo \citep{GFSAD, CopernicusLandCover} while also being higher resolution and more recent. We have delivered this product to the Togolese government where it is being used to help inform decision-making in response to the COVID-19 pandemic, thus demonstrating a successful transition of machine learning research to operational rapid response for a real humanitarian crisis. We have made our map as well as all data and code used in this study publicly available to facilitate future research and operational systems in data-sparse regions. 
\begin{figure}
  \centering
  \includegraphics[width=0.75\linewidth]{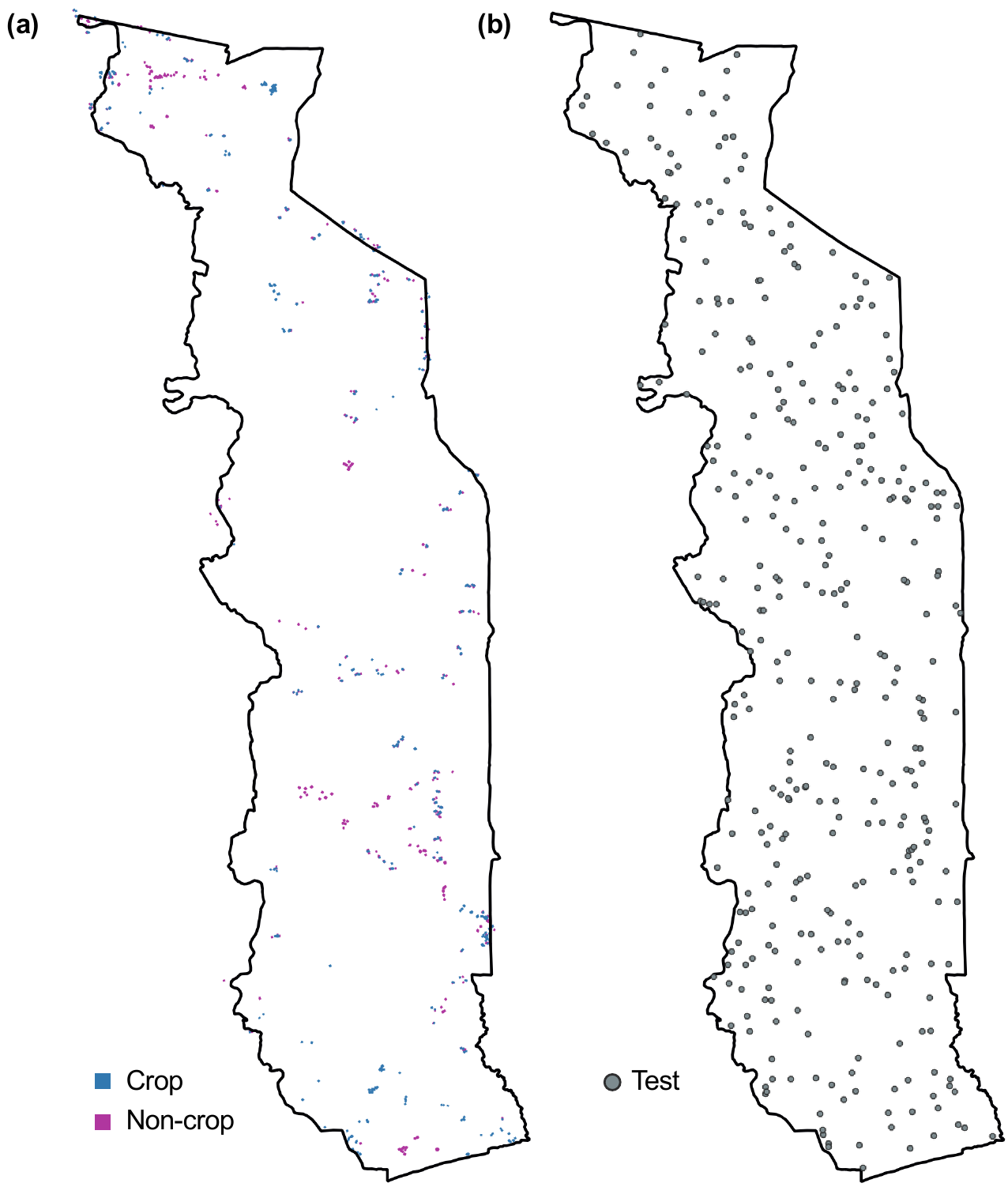}
  \caption{Distribution of labeled examples in training (a) and test (b) datasets.}
  \label{fig:labels}
\end{figure}

\section{Methods}

\subsection{Labeled Data}
\subsubsection{Crowdsourced Labels}
We leveraged a large (35,866 labels) crowd-sourced dataset with labels of ``crop'' or ``non-crop'' from diverse, globally-distributed locations \citep{GeoWiki} to help the model learn features that are useful for detecting a wide range of cropland globally. Details about sample selection, labeling procedures, and quality assessment/control measures can be found in \citet{Bayas2017}. Each location was labelled by multiple labelers; to turn these labels into binary labels, we took the mean of the labels for each location, with a threshold of 0.5 to define a point as ``crop'' or ``non-crop.''

\subsubsection{Active Labeling}
To generate training labels, experts in photointerpretation of agricultural land cover (authors Becker-Reshef, Barker, Hosseini, Kerner, Munshell, and Paliyam) drew polygons over pixels with labels of ``crop'' or ``non-crop.'' We used a combination of basemaps for photointerpretation: a SkySat 72cm composite of January-March 2019 \citep{Planet}, PlanetScope composites of April-July and July-October 2019 \citep{Planet}, and Google Earth Pro basemaps (comprised primarily of high-resolution Maxar images). We used QGIS and Google Earth Pro to draw polygons over basemaps. The locations of labels were chosen to be distributed across different agroecological zones and non-crop land cover types observed across Togo. The central pixel within each polygon was used for training. We obtained 394 crop and 194 non-crop labels for the initial training dataset to supplement the 43 examples within Togo in the \citet{GeoWiki} dataset (631 examples total). After training the model with the hand-labeled and \citet{GeoWiki} dataset, we collected additional labeled examples (294 crop and 394 non-crop) by analyzing the predicted map and identifying areas of model confusion (e.g., transient vegetation on shorelines that the model might confuse with crops). This allowed us to maximize the utility of each training example, as it focused on places where the model was weakest rather than adding redundancy to the dataset. This resulted in a final training dataset of 1,319 hand-labeled examples, visualized spatially in Figure \ref{fig:labels}a. 

\subsection{EO Data}
We used Copernicus Sentinel-2 surface reflectance (Level 2A) observations that corresponded to label locations as input to the model. We exported 160 m $\times$ 160 m patches for each labeled pixel using Google Earth Engine (GEE) \citep{GORELICK201718}, then extracted the closest pixel within the patch to the label location. To construct a cloud-free timeseries representation of each pixel, we used the algorithm from \citet{CloudFreeSentinel}, which finds the least cloudy pixel within a defined time period. We used 12 30-day time periods to construct a monthly timeseries spanning 360 days for each labeled pixel. We used all spectral bands except for B1 (coastal aerosol) and B10 (cirrus SWIR), and added the normalized difference vegetation index (NDVI), computed as $\text{NDVI} = \frac{\text{B08}-\text{B04}}{\text{B08}+\text{B04}}$. All bands (which range from 10 to 60m resolution) were upsampled to 10 m during GEE export. We used observations acquired between March of Year N and March of Year N+1 where N is the year the labels were created for. Thus for \citet{GeoWiki}, we used observations acquired March 2017-March 2018 and for our hand-labeled dataset we used observations acquired March 2019-March 2020.

\subsection{Model} \label{section:model}
\begin{figure}
  \centering
  \includegraphics[width=\linewidth]{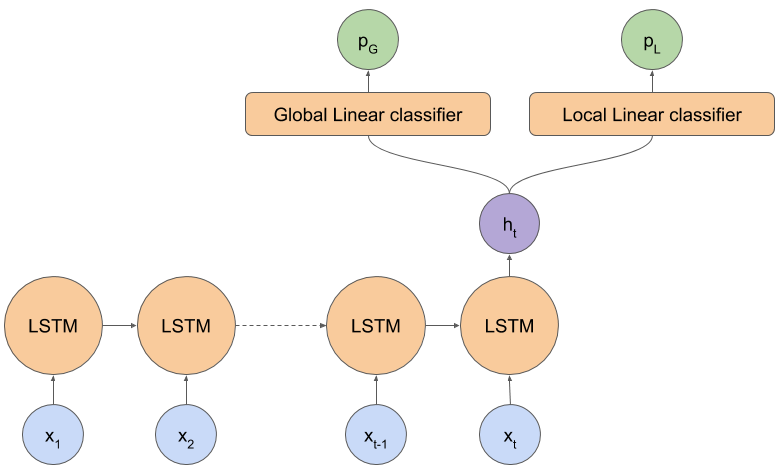}
  \caption{A multi-headed LSTM based model for pixel-wise crop identification. The model has two classification layers: a global linear classifier to classify instances outside Togo, and a local linear classifier to classify instances within Togo.}
  \label{fig:lstm}
\end{figure}

We trained a one layer multi-headed LSTM model (Figure \ref{fig:lstm}) to predict whether a pixel contained cropland (1) or not (0). The input was expressed as a timeseries $\bm{X} = \{\bm{x}_{1}, \bm{x}_{2}, ... \bm{x}_{12}\}$, where each timestep $\bm{x}_{i} \in \mathbb{R}^{1\times14}$ consisted of the least-cloudy composite over a 30-day period of 11 optical bands (all bands except B01 and B10) plus NDVI. The LSTM hidden layer had 64 units. A classifier and a sigmoid activation were used on the final hidden output to map the hidden layer output (logits) to a value that can be interpreted as the posterior probability of crop presence in the pixel.

We performed a grid search that optimized AUC ROC on the validation set to choose values for two hyperparameters: number of linear layers in the classifier (1 or 2) and whether or not a dropout of 0.2 should be applied between each timestep in the LSTM. 

To allow the model to focus on examples within Togo while also learning from the global \citet{GeoWiki} dataset, we employed a multi-headed LSTM model, where one classifier was used for all examples within Togo (local instances) and another was used for all other examples (global instances). All instances were shuffled during training: a batch could therefore contain a combination of global and local data instances. To combine the results from both classifiers, we used the following loss function to train the model:
\begin{equation}
\mathcal{L} = \frac{W}{\alpha}\mathcal{L}_{\textrm{global}} + \mathcal{L}_{\textrm{local}}
\end{equation}
Where $\mathcal{L}_{\textrm{global}}$ and $\mathcal{L}_{\textrm{local}}$ are the binary cross-entropy losses for the global and local data instances respectively, and $\alpha$ is a weighting parameter. $W$ is a batch-specific value, used to weight the ratio of global and local instances in a batch:
\begin{equation}
    W = \frac{\mbox{Number of global instances in batch}}{\mbox{Number of local instances in batch}}
\end{equation}
In our experiments, we used $\alpha = 10$. We used the Adam optimizer with default parameters to tune weights during training \citep{Kingma2014}. To determine when to stop training, we used early stopping with a patience of 10 using a validation set consisting of 20\% of the combined \citet{GeoWiki} and hand-labeled Togo datasets. We implemented the model and experiments using PyTorch \citep{pytorch} in python. All code is publicly available at \url{https://github.com/nasaharvest/togo-crop-mask}.

\begin{table*}
  \caption{Results for each method evaluated on the test set. The best results are in bold.}
  \label{tab:experimental-results}
  \begin{tabular}{cccccc}
    \toprule
    Description &Accuracy &AUC &Precision &Recall &F1\\
    \midrule
    \citet{CopernicusLandCover} (Copernicus 100m) & 0.67 & 0.74 & 0.54 & 0.24 & 0.33\\
    Single-headed LSTM (hand-labeled only) & 0.71 & 0.90 & 0.59 & \textbf{0.93} & 0.72\\
    \citet{GFSAD} (GFSAD 30m) & 0.74 & - & 0.62 & 0.62 & 0.62 \\
    Support Vector Machine (hand-labeled only) & 0.79 & 0.88 & 0.65 & 0.84 & 0.73 \\
    Random Forest (hand-labeled only) & 0.81 & 0.90 & 0.66 & 0.88 & \textbf{0.76} \\
    Multi-headed LSTM (hand-labeled + \citet{GeoWiki}) & \textbf{0.83} & \textbf{0.91} & \textbf{0.81} & 0.68 & 0.74\\
    \bottomrule
  \end{tabular}
  \label{experimental-setup-results}
\end{table*}

\section{Results}
\label{sec:results}

We selected points for a test dataset by randomly sampling 350 points within the Togo country boundaries, constrained by a minimum distance of 50 m between samples (Figure \ref{fig:labels}b). Four experts (authors Barker, Kerner, Munshell, and Nakalembe) labeled each example. Labels were determined by majority vote in order to mitigate label noise in the absence of groundtruth validation, and ties were discarded. The consensus test dataset contained 106 crop and 200 non-crop examples (306 total test examples).

We compared two experimental setups to evaluate the contribution of the \citet{GeoWiki} dataset:
\begin{itemize}
    \item A single-headed LSTM trained on only the hand-labeled Togo data
    \item A multi-headed LSTM trained on both the hand-labeled Togo and \citet{GeoWiki} data 
\end{itemize}

In both cases, we used a model with hyperparameters determined by the grid search described in \ref{section:model}. For the single headed model, this consisted of a single classification layer in the classifier \textit{without} dropout in the LSTM. For the multi-headed model, this consisted of two classification layers in the classifier \textit{with} dropout in the LSTM.

\begin{figure}
  \centering
  \includegraphics[width=0.9\linewidth]{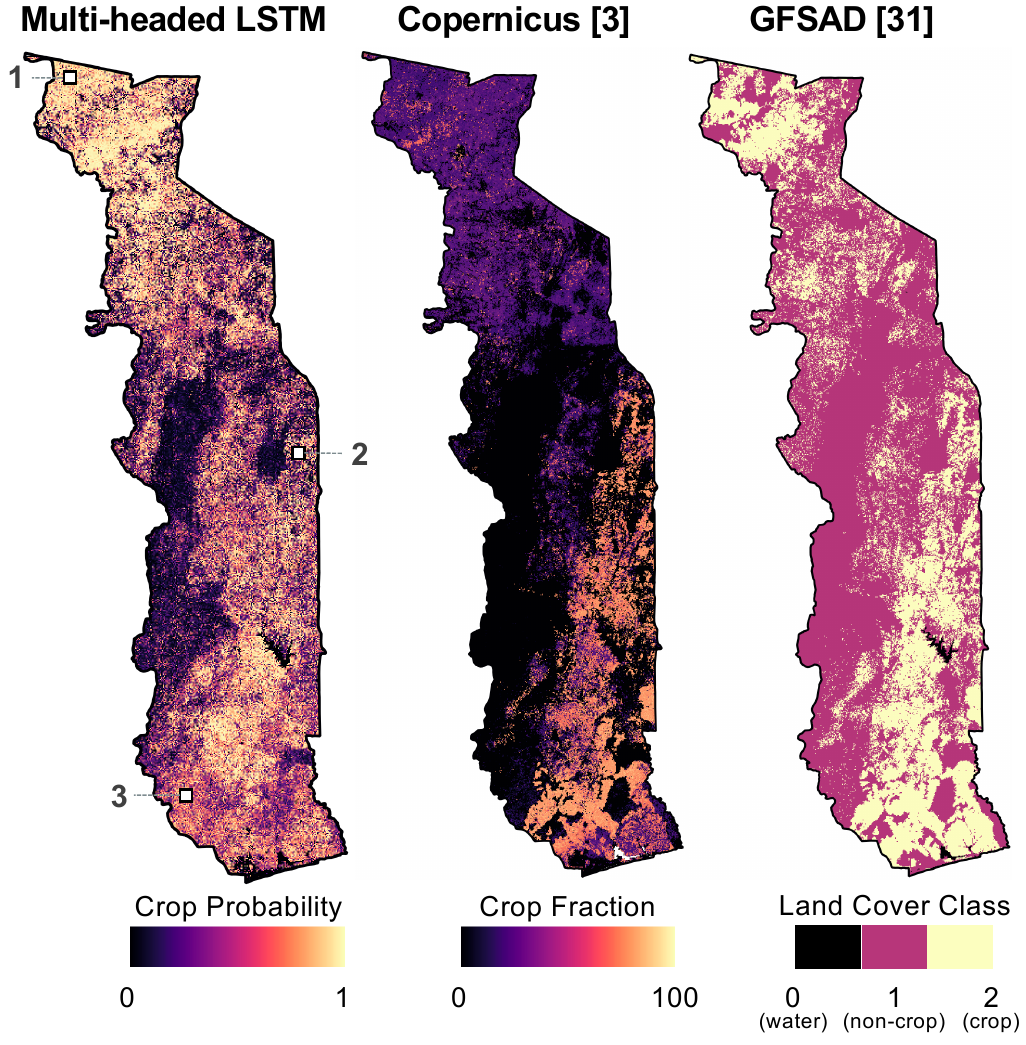}
  \caption{Cropland maps of Togo generated using our method, \citet{CopernicusLandCover}, and \citet{GFSAD}. White boxes indicate locations of insets shown in Figure \ref{fig:comparisons}.}
  \label{fig:fullmap}
\end{figure}

\begin{figure*}
  \centering
  \includegraphics[width=0.9\linewidth]{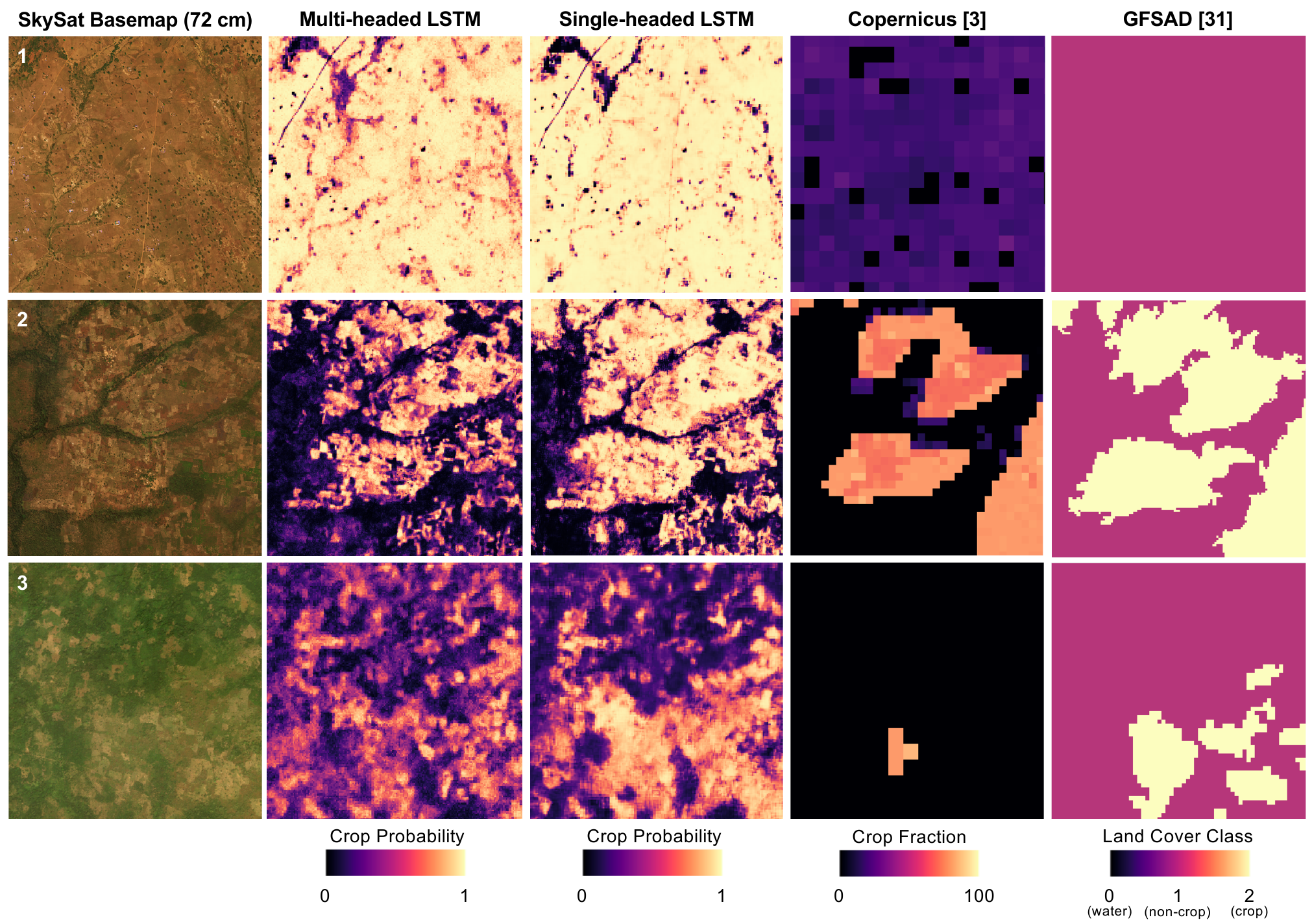}
  \caption{Qualitative comparison of cropland maps produced using our multi-headed and single-headed LSTM methods, \citet{CopernicusLandCover}, and \citet{GFSAD} for three example regions, with SkySat 72 cm basemap (April-March composite) for reference. Locations of each inset indicated by white boxes in Figure \ref{fig:fullmap}.}
  \label{fig:comparisons}
\end{figure*}

We compared our results to two publicly-available cropland maps of Togo. One map is Global Food Security-support Analysis Data (GFSAD) cropland extent map of Africa \citep{GFSAD}. This map has 30 m resolution and was produced using two pixel-based supervised classifiers, a Random Forest and a Support Vector Machine (SVM), and one object-based classifier, a Recursive Hierarchical Image Segmentation classifier. The GFSAD map does not provide cropland probabilities, but a discrete class assigned to each pixel of 0 (water), 1 (non-cropland), or 2 (cropland). The most recent GFSAD product is from 2015. Prior to our map, the GFSAD map is to our knowledge the highest resolution cropland map publicly available for Togo. Figure \ref{fig:fullmap} (right) shows the GFSAD land cover map for Togo. The second map we compared our results to is the Copernicus Land Cover cropland map \citep{CopernicusLandCover}. The Copernicus map specifies the fraction of each pixel covered by crops (using values 0-100) at 100 m resolution and was produced using a combination of random forests and expert rules with satellite data inputs from the PROBA-V sensor. The most recent Copernicus map is for 2018. To enable quantitative comparison with our results, we treated crop fraction as analogous to crop probability and computed all performance metrics using our test dataset (Table \ref{tab:experimental-results}); however, we emphasize that this may not be how the map was intended to be used and qualitative comparisons may be more faithful to the intended use of the \citet{CopernicusLandCover} map. Figure \ref{fig:fullmap} (center) shows the Copernicus crop fraction map for Togo. A direct comparison between our method and \citet{GFSAD} and \citet{CopernicusLandCover} would require running all three methods on the same data (i.e., 10m Sentinel-2 observations paired with our labeled examples). Since neither the code for these methods nor sufficient information for reproducing them faithfully are available, we additionally performed two baseline comparisons to Random Forest and SVM classifiers which are the basis for the \citet{GFSAD} and \citet{CopernicusLandCover} methods.
For the Random Forest, we used 100 estimators with no maximum depth (nodes were expanded until all leaves were pure, or until the leaves contained fewer than 2 samples). For the SVM, we used a radial basis function kernel. We implemented both methods using Scikit-learn in python \citep{scikit-learn}.

To assess performance, we computed the area under the receiver operating characteristic (ROC) curve (AUC score), accuracy (0.5 threshold on posterior probabilities), precision (user's accuracy), recall (producer's accuracy), and F1-score (harmonic mean between precision and recall). We did not compute AUC for GFSAD, which only provides the discrete predicted class. The results are reported in Table \ref{tab:experimental-results}. Figure \ref{fig:fullmap} (left) shows our predicted cropland map.

To qualitatively compare our method compared with \citet{GFSAD} and \citet{CopernicusLandCover}, we show each map compared to the SkySat 72cm basemap (true color) for several example locations in Figure \ref{fig:comparisons}. Each region in Figure \ref{fig:comparisons} contains primarily cropland, though the level of vegetation in non-crop areas varies in each image since Togo's climate ranges from dry savanna in the north to tropical in the south. While all maps detect most fields in the second row, the \citet{GFSAD} and \citet{CopernicusLandCover} maps failed to detect most of the fields in the first and third rows.

Our resulting map of cropland probabilities at 10 m resolution as well as all hand-labeled training and testing labels are publicly available at: \url{https://doi.org/10.5281/zenodo.3836629}.

\section{Discussion}
 The multi-headed model trained on the combined hand-labeled and \citet{GeoWiki} dataset had the highest score for most performance metrics. While AUC score is a measure of model performance across a range of possible thresholds on posterior probability, the decision boundary when using the sigmoid function as the output activation layer is 0.5. While in practice a different threshold could be chosen, it is not clear how to construct a large and diverse enough validation set that could be used to reliably select an alternate threshold over very large, heterogeneous regions (e.g., the country of Togo). Thus, the improved accuracy and precision of the multi-headed model (using a threshold of 0.5) - in addition to its improved AUC score - made it preferable. Additionally, since the multi-headed model included training data from prior years (2017) and global instances of cropland \citep{GeoWiki}, we expect that this model would generalize better to potential domain shift in future years than the single-headed model trained with only one year of local data. Figure \ref{fig:comparisons} shows that the multi-headed LSTM and single-headed LSTM have similar patterns of detection, but overall the multi-headed LSTM has less extreme probabilities (values closer to 0.5 than 0 or 1). This suggests that the global examples used to train the multi-headed LSTM act like a regularizer for the classifier by moderating the confidence of predictions and preventing overfitting.
 
 We found that the \citet{CopernicusLandCover} map had the lowest performance for the test dataset, with an F1 score significantly lower than other methods, though we note that these results could underestimate the performance because the map reports the fraction of each pixel covered by crops, not probability, in each pixel. This lower performance may, in part, be due to the coarse spatial resolution of EO data used to produce the map and the growth in cropland area that has likely occurred since the map was produced in 2015. Similarly, the lower performance of the \citet{GFSAD} could also, in part, be due to lower spatial resolution of input data (30 m) and changes in cropland area that might have occurred between 2018 (when the most recent \citet{GFSAD} map was produced) and 2019.
 
 \subsection{Sensitivity to number of local examples}

\begin{figure}
    \centering
    \includegraphics[width=0.9\linewidth]{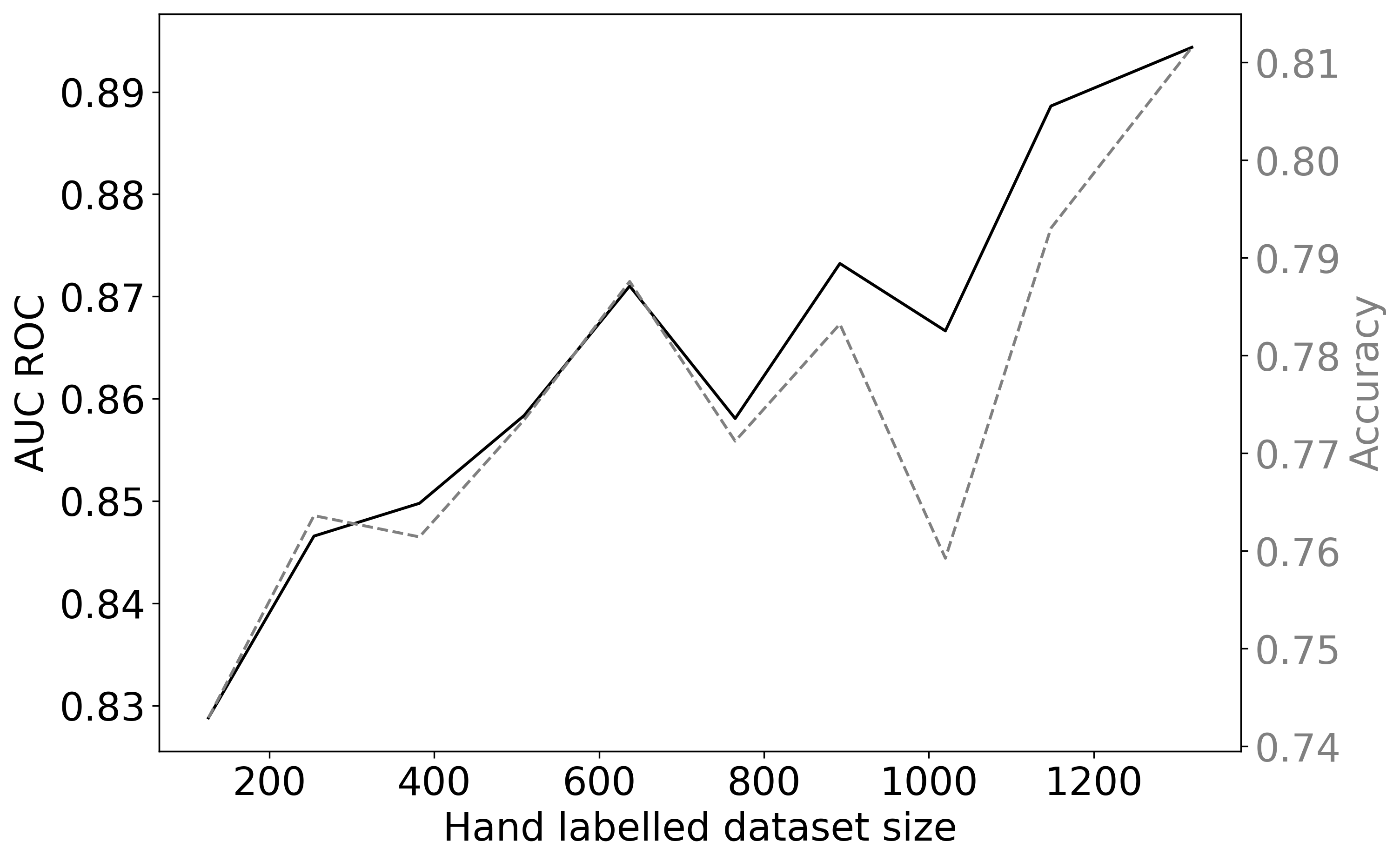}
    \caption{A plot of AUC ROC and accuracy scores of the multi-headed model as the size of the hand-labeled dataset size increases. Each score is calculated from a mean of 3 runs with different random seeds.}
    \label{fig:auc_from_size}
\end{figure}

To assess the sensitivity of the model to the frequency and number of local (Togo) examples in the training dataset and provide a guide for future labeling campaigns, we measured the performance of the multi-headed model on the test set as a function of the number of hand-labeled examples used in training (Figure \ref{fig:auc_from_size}). Although the performance of the model increased as the size of the hand-labeled dataset increased, the model performs well even with very little labeled data. This suggests this technique may be applied to other data-sparse regions with a comparable or fewer number of labels as we created for this study.
 
\subsection{Consensus in photointerpreted labels}

\begin{figure}
  \centering
  \includegraphics[width=0.7\linewidth]{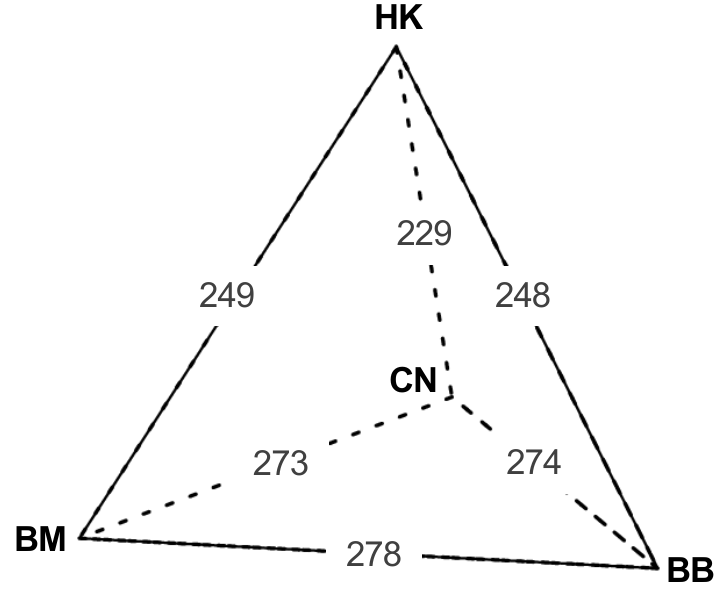}
  \caption{Tetrahedron showing number of examples (of 350 total) for which each pair of labelers (indicated by initials) agreed.}
  \label{fig:consensus}
\end{figure}

As described in Section \ref{sec:results}, we created a test dataset by randomly sampling 350 points within Togo and having four experts label each point as ``crop'' or ``non-crop'' using photointerpretation of high-resolution satellite imagery. For each point, we chose the label assigned by the majority of labelers (ties were discarded). Figure \ref{fig:consensus} shows the number of examples for which each pair of labelers chose the same label. Agreement between pairs of labelers ranged from 229-278 (65\%-79\%) of the total points and all labelers agreed unanimously on 181 (52\%) of the points. This level of consensus was surprisingly low, and while our photointerpretation-based dataset is a high-quality surrogate for ground-truth labels, it underscores the difficulty for even experts to determine land cover types from high-resolution satellite imagery, particularly in developing countries. This should be considered in the design of future photointerpretation-based labeling efforts, especially crowdsourced efforts in which labelers may not be experts. We plan to conduct a future study to assess agreement between labels determined by experts for the same points from photointerpretation vs. field observation to better characterize the level of accuracy to expect from photointerpreted labels for different regions and land cover types.

\subsection{Lessons learned for rapid response}
The urgent need for a cropland map of Togo that captured smallholder farms (usually $<$1 ha) required us to leverage our research methods to rapidly deliver a result for operational use in less than 10 days. During this process, we gathered several ``lessons learned'' that we share to help prepare other researchers to leverage their methods for rapid humanitarian response in the future, as well as encourage community practices that facilitate rapid response.
\begin{itemize}
\item Good software engineering practices (such as developing modular and well-documented code) is often prioritized in operational rather than research settings. However, developing research methods motivated by their use for rapid humanitarian response in the future requires integration of operational practices into research workflows. We found that developing research methods using operational software engineering practices enabled us to quickly execute experiments and apply the method to large geographic areas when the need for rapid response arose.
\item Use of cloud services (in this case AWS) allowed us to easily scale our model (particularly during country-wide inference) and enable faster computation in a short period of time. Additionally, the use of EC2 instances enabled processes to be run, monitored, and modified by a distributed team rather than a single user.
\item File transfer/download remains a bottleneck for rapid response workflows. Transferring data from storage (e.g., via Google Earth Engine, the Sentinel-2 AWS S3 bucket, or USGS Earth Explorer) to compute (e.g., AWS) constituted a significant part of the time to deliver our cropland map. There is a community need for better tools to enable searching, querying, and subsetting of geospatial data on cloud platforms.
\item Given the complexity of deep learning architectures such as Inception-v3 \citep{Szegedy2016} or ResNet \citep{He2016} designed for image classification, it might seem that classification of land cover or crops in remote sensing data---which often exhibit high intraclass variance and low inter-class variance---would also require complex architectures. However, we found that a relatively simple, shallow architecture achieved good results for this task, and has the benefit of being faster in training/inference, requiring fewer examples, and being less susceptible to overfitting than more complex architectures.
\item The GFSAD \citep{GFSAD} and Copernicus \citep{CopernicusLandCover} cropland maps were easy to find and download and were well-documented, which enabled us to quickly benchmark our map against existing maps. To give others this same benefit with our cropland map, we have made our cropland map, code, and training labels publicly available.
\end{itemize}

\section{Conclusion}
Cropland maps are critical inputs for decision-makers to rapidly design and enact policies during humanitarian crises, e.g., to deliver aid or boost productivity for farmers, yet unavailability of high-resolution, up-to-date cropland maps in most countries globally precludes their use for rapid response. The development of cropland maps, in turn, is limited by the availability of ground data that can be used for training and validating machine learning classifiers. 

We present a method for cropland classification in regions with little to no available ground data that uses a multi-headed LSTM network to learn global and local features for identifying cropland based on multispectral time-series Sentinel-2 observations. We used this method to create a cropland map of Togo at 10 m resolution in fewer than 10 days to assist the Togolese government in decision-making about aid distribution to farmers during the COVID-19 crisis. This demonstrated the successful transition of machine learning research to operational use in rapid response for a real humanitarian crisis, and we provided lessons learned from this experience to facilitate this transition for other researchers in the future. Additionally, our approach gives an example of how to leverage existing readily-available labeled datasets and reduce the number of data points to be collected in field campaigns, which can be costly as well as infeasible due to regional insecurity or travel restrictions (e.g., during the COVID-19 pandemic). In future work, we plan to investigate leveraging our multi-headed LSTM approach for crop type classification, subsetting the Geo-Wiki dataset based on geographical or agroecological zones, and including Sentinel-1 synethic aperture radar data in the model input to mitigate the effects of clouds during the growing season. Additionally, we will apply our method for cropland mapping in other data-sparse African countries.

\begin{acks}
We would like to thank Tanya Harrison, Zara Khan, and Charlie Candy from Planet, Inc., for making SkySat and PlanetScope basemaps available for this study. We are also grateful to Noel Gorelick from Google Earth Engine for helping to accelerate our Google Earth Engine processing workflow in order to meet this rapid response timeline. Finally, we would like to thank Steffen Fritz, Linda See, and the rest of the Geo-Wiki team for providing the publicly-available, global, crowdsourced cropland dataset critical to our method.
\end{acks}

\bibliographystyle{ACM-Reference-Format}
\bibliography{bib}

\end{document}